\documentclass{article}
\usepackage{spconf,amsmath,graphicx}

\graphicspath{{./}}
\usepackage{amssymb}
\usepackage{color}
\usepackage{bm}
\usepackage{siunitx}
\usepackage{url}

\title{
  Transfer learning from synthetic to real images using variational autoencoders for precise position detection\thanks{\copyright 2018 IEEE --- Manuscript accepted at ICIP 2018.}
}
\name{Tadanobu Inoue$^{1}$, Subhajit Chaudhury$^{1}$, Giovanni De Magistris$^{1}$ and Sakyasingha Dasgupta$^{2\dagger}$
}
\address{
  $^{1}$IBM Research, Japan. \{inouet, subhajit, giovadem\}@jp.ibm.com \\
  $^{2}$Ascent Robotics Inc., Japan, sakya@ascent.ai
  \thanks{$^{\dagger}$ This work was carried out during his position with IBM Research.}
}

\begin{document}
%
\maketitle
\begin{abstract}
Capturing and labeling camera images in the real world is an expensive task, whereas synthesizing labeled images in a simulation environment is easy for collecting large-scale image data.
However, learning from only synthetic images may not achieve the desired performance in the real world due to a gap between synthetic and real images.
We propose a method that transfers learned detection of an object position from a simulation environment to the real world.
This method uses only a significantly limited dataset of real images while leveraging a large dataset of synthetic images using variational autoencoders.
Additionally, the proposed method consistently performed well in different lighting conditions, in the presence of other distractor objects, and on different backgrounds.
Experimental results showed that it achieved accuracy of \SI{1.5}{\milli\metre} to \SI{3.5}{\milli\metre} on average.
Furthermore, we showed how the method can be used in a real-world scenario like a ``pick-and-place'' robotic task.
\end{abstract}
\begin{keywords}
deep learning, position detection, transfer learning, variational autoencoder, computer simulation
\end{keywords}

\section{Introduction}
\label{sec:intro}
Supervised deep learning tasks require a large collection of labeled data for producing generalizable performances of unseen test data, and in estimating the location of objects~\cite{Leitner2013, Collet2010, Tang2012}.
For image-based learning, it is time-consuming to capture and label camera images in the real world.
In contrast, it is easy to synthesize and collect large-scale labeled images in a simulation environment.
Since there is a ``reality gap'' between simulation and real environments, it is difficult to match performances in the real world by learning only from these synthesized images.
Thus, there is a need to bridge the gap between real and simulated images to learn useful features in a cross-domain manner.

\begin{figure}[thpb]
	\centering
	\includegraphics[width=0.99\linewidth]{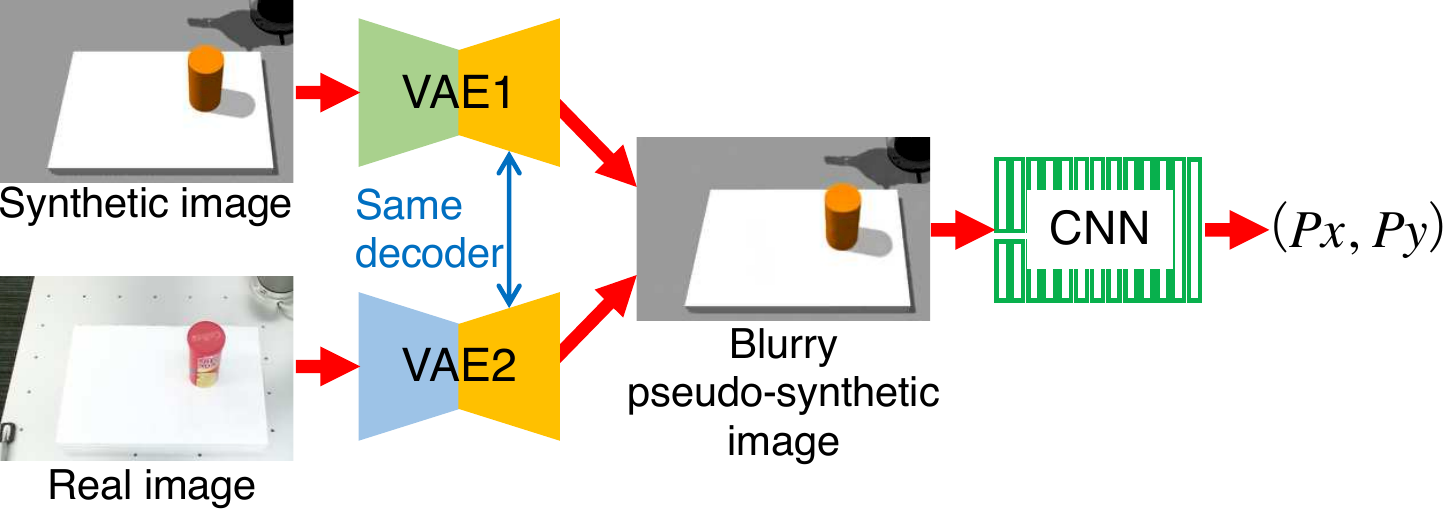}
	\caption{Proposed concept for detecting object position}
	\label{fig:concept}
\end{figure}

To overcome this gap, Shrivastava \textit{et al.}~\cite{Shrivastava} proposed a method to generate realistic images by refining synthesized ones using adversarial deep learning.
Santana and Hotz~\cite{Santana} combined variational autoencoders (VAE)~\cite{KingmaICLR2014} and generative adversarial networks (GAN)~\cite{goodfellow2014generative} to generate realistic road images.
To generate these realistic images, the approaches needed numerous unlabeled real images for adversarial training during refinement.
However, collecting a large number of unlabeled real images can be extremely time-consuming and difficult in some cases.
For example, it might be cumbersome to capture all possible combinations of object types and locations in a given scene configuration.

Domain randomization approaches ~\cite{Sadeghi2017, TobinIROS2017, Yan2017, Zhang2017} provide promising methods to overcome the reality-gap by training multiple random configurations in simulations without many real images.
It was argued that if the network was trained on multiple random configurations, a real image could also be treated as yet another random configuration.
The literature also shows examples, where domain adaptation methods have been used for wider robotic applications~\cite{Rusu, Higgins, Gupta2017, James, Peng2017} beyond vision-based tasks~\cite{Peng2015, Kulis2011}.

We propose a transfer learning method for detecting real object positions from RGB-D image data using two generative models that generate common pseudo-synthetic images from synthetic and real images.
This method uses a significantly limited dataset of real images, which are typically costly to collect while leveraging a large dataset of synthetic images that can be easily generated in a simulation environment.
Furthermore, the proposed model remains invariant to changes in lighting conditions, the presence of other distractor objects, or backgrounds.
The obtained precision in detecting the position of the desired object ensures real-world application potential.
We demonstrate its application in a typical robotic ``pick-and-place'' task as shown in the video (\url{https://youtu.be/30vji7nJibA}).

\section{DETECTING OBJECT POSITIONS USING VAEs}
\label{sec:method}

Figure~\ref{fig:concept} depicts the concept of our proposed method.
The core idea of this work is that the distribution of image features may vary between simulated and real environments, but the output label of the object position should remain invariant for the same scene.

\begin{figure}[thpb]
  \centering
  \includegraphics[width=0.98\linewidth]{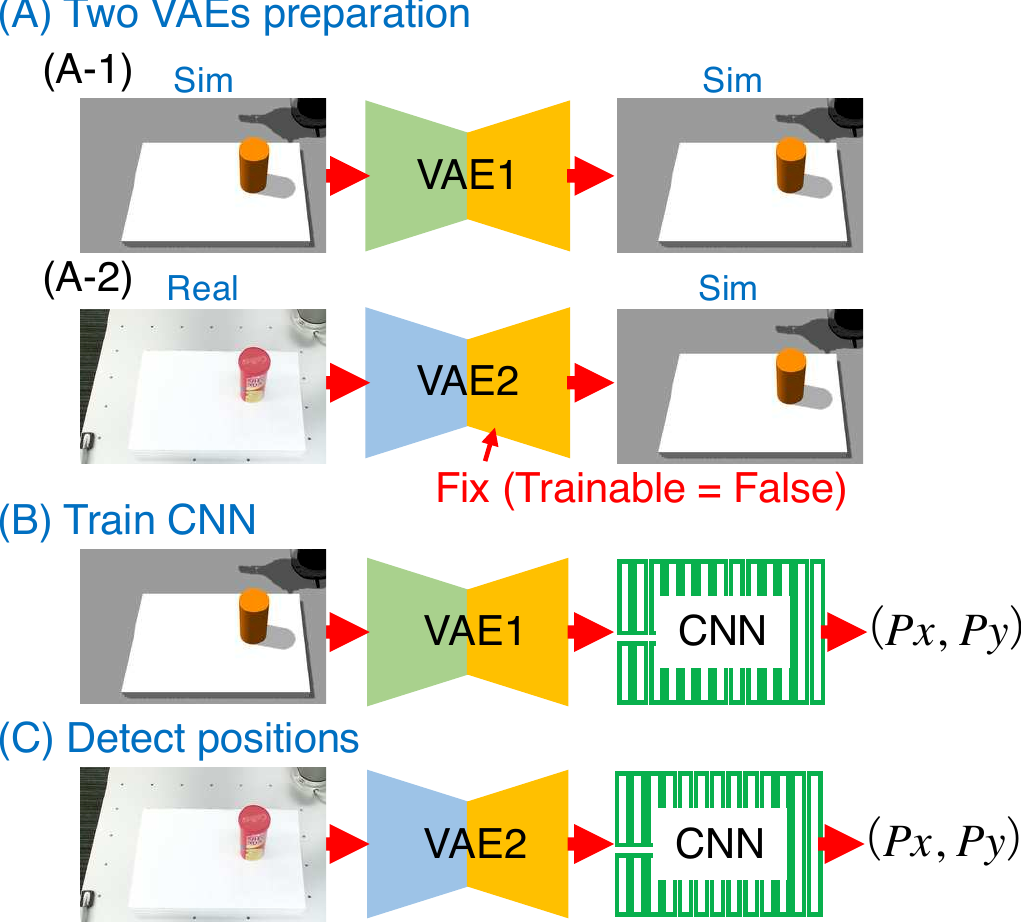}
  \caption{Three steps of our proposed method: (A) Train two VAEs sequentially; (A-1) Train VAE1 to generate synthetic images; (A-2) Train VAE2 to generate synthetic images from real images; (B) Train CNN to detect object position using VAE1 outputs with synthetic images; (C) Detect real object positions using VAE2 outputs with real images and CNN}
  \label{fig:three_steps}
\end{figure}

Our method consists of three broad steps as shown in Fig.~\ref{fig:three_steps}.
We use two VAEs for generating similar common images from synthetic and real image data and use this common image data to train a convolutional neural network (CNN) for predicting the object position with improved accuracy.
Here, although two VAEs have distinct encoder layers as generative models for images, they have the same decoder, which is used to train the CNN.
Thus, even if the VAE generates blurry images, the CNN will learn to predict object positions from this skewed but common image space.
Furthermore, since the CNN can be trained with many generated images from the synthetic domain, we can achieve improved object position estimation from a significantly limited set of labeled real images.

\subsection{Variational real to synthetic mapping}
Two VAEs are prepared to generate common pseudo-synthetic images from synthetic and real images.
A simulation environment is set up and large-scale synthetic images are captured along with corresponding ground-truth object position labels.
We train VAE1, which encodes and decodes from a synthetic image to the same synthetic image as shown in Fig.~\ref{fig:three_steps} (A-1).

The weights from VAE1 is used to initialize another VAE network with the same structure (VAE2).
This VAE learns a conditional distribution that encodes and decodes from a real image to the corresponding synthetic image as shown in Fig.~\ref{fig:three_steps} (A-2).
During the training, we fix the decoder layers and adapt only the parameters for the encoder, which receives the real images as input.
This is equivalent to forcing the latent space obtained from the synthetic and real images to be identical.
The learned encoder and decoder can be combined to generate pseudo-synthetic images as output from the corresponding real image as input.

\subsection{Object detection on common image space}
A CNN is trained to detect object positions as shown in Fig.~\ref{fig:three_steps} (B).
To close the gap between synthetic and real images, we use the outputs of the trained VAE1 in the previous step, instead of using synthetic images directly.
Since both VAE1 and VAE2 use the same decoder (generator), the image space of both the outputs is the same which enables cross-domain transfer of learned tasks.
This forms the primary idea that is presented in this paper.
Due to the availability of a large training dataset synthesized in a simulation environment, we can train the CNN adequately to obtain an accurate object detector.

Finally, in the test phase, object positions are detected in the real world as shown in Fig.~\ref{fig:three_steps} (C).
In this case, VAE2 outputs blurry pseudo-synthetic common images, and the CNN trained with the similar common images outputs the object position.

\section{EXPERIMENTS}
\label{sec:experiments}

\subsection{Experimental setup}

For all experiments, we used a Gazebo\textsuperscript{\textregistered}~\cite{Gazebo} simulation environment and Kinect\textsuperscript{\textregistered}~\cite{Kinect} camera.
In Gazebo, object models were located on a white styrofoam ($45\times30$ \SI{}{\centi\metre}) at specific positions.
A corresponding Kinect model was also loaded in Gazebo to capture RGB-D images of the workspace scene.
At the same time, objects were manually located on a same-sized white styrofoam at the specific positions in the real world.
Furthermore, the images captured in both Gazebo and the real world were cropped to a smaller region.

\begin{figure}[thpb]
  \centering
  \includegraphics[width=0.98\linewidth]{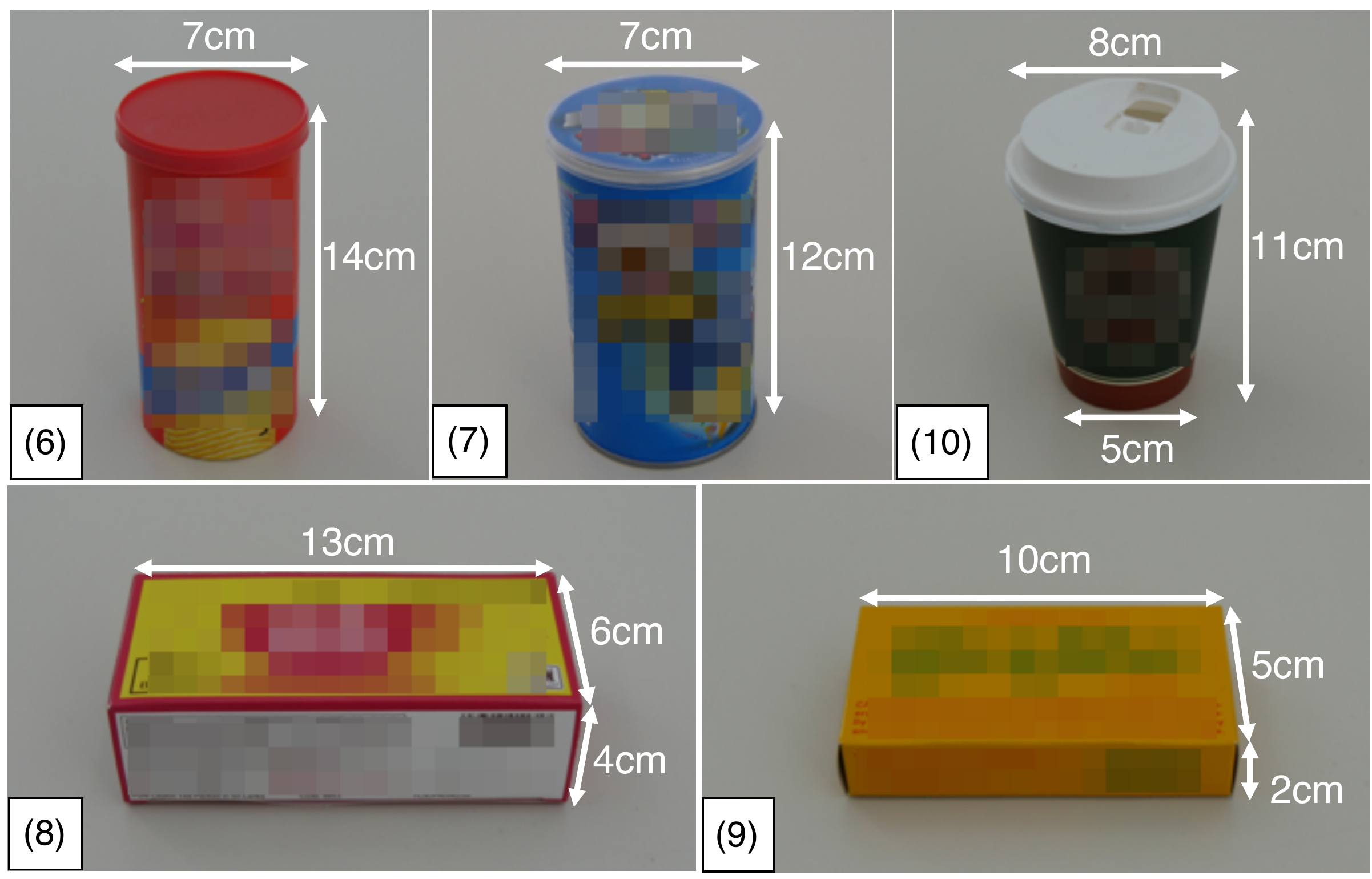}
  \caption{Complex textured household objects}
  \label{fig:natural_obj}
\end{figure}

Our method was evaluated in $14$ experiments (a)-(n).
We used five simple real objects created by a 3D printer as shown in Table~\ref{tb:test_obj}, for experiments (a)-(g) and five complex textured household objects as shown in Fig.~\ref{fig:natural_obj}, for experiments (h)-(n).

\vspace{-0.0cm}
\begin{table}[thbp]
\caption{Simple real objects}
\label{tb:test_obj}
\begin{center}
\begin{tabular}{|c|c|c|c|c|}
\hline
No. & Experiments & Color & Shape & Size (\SI{}{\centi\metre}) \\
\hline
(1) & (a), (b) & red & cube & $5\times5\times5$ \\
\hline
(2) & (c), (f), (g) & green & cube & $4\times4\times4$ \\
\hline
(3) & (d), (g) & black & cylinder & \shortstack{radius $3.5$ \\ height $1$ } \\
\hline
(4) & (e), (g) & blue & \shortstack{triangular \\ prism} & \shortstack{radius $4.5$ \\ height $1$ } \\
\hline
(5) & (g) & red & cube & $4\times4\times4$ \\
\hline
\end{tabular}
\end{center}
\end{table}

\subsection{Evaluation on simple real objects}
\label{sec:simple_obj}

First, as a baseline experiment (a), we evaluated the position detection of a red cube (1) in a naive manner, using the CNN trained with $54$ real images at \SI{5}{\centi\metre} grid positions directly.
Then, our method was applied to the red cube (1), green cube (2), black cylinder (3), and blue triangular prism (4) (experiments (b)-(e)).
We used $4131$ synthetic images in which each synthetic object was located at \SI{5}{\milli\metre} grid positions for training VAE1.
Only $54$ real and $54$ synthetic images in which each object was located at \SI{5}{\centi\metre} grid positions for training the VAE2 were used.
This real image data was not augmented~\cite{Bateux}.

In experiment (b), the mean squared error (MSE) of synthetic and real input images was $9.9 \times 10^3$ on average and the MSE of two VAE output images was $2.7 \times 10^{-3}$ on average.
Our method using VAEs improved the similarity between images in real and synthetic domains for the inputs of later CNNs, which then overcame the ``reality gap.''

\begin{figure}[thpb]
  \centering
  \includegraphics[width=0.98\linewidth]{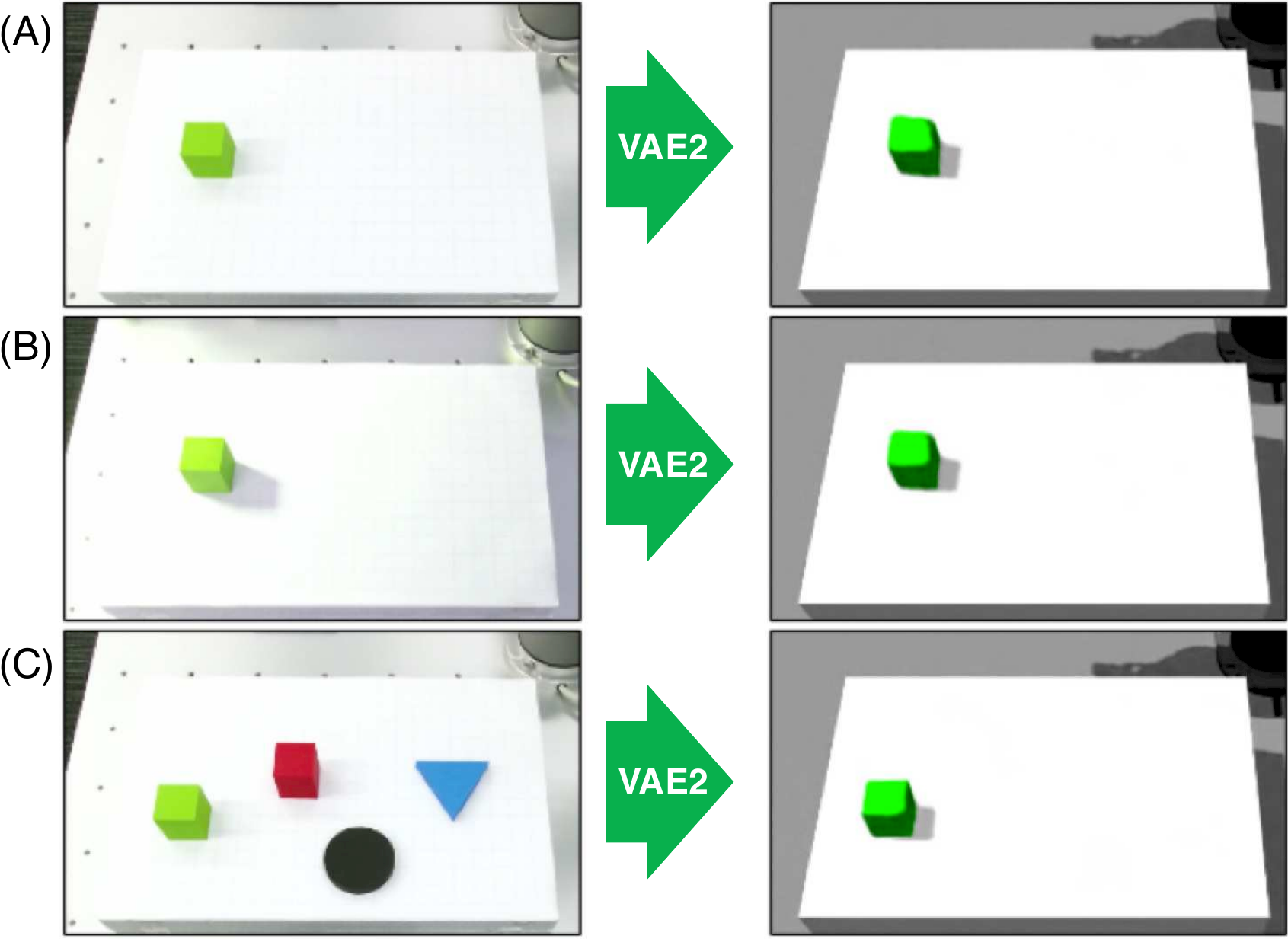}
  \caption{Images generated by VAE2 under different lighting conditions and with the presence of distractor objects: (A) Room light turned on and table light turned off; (B) Room light turned off and table light turned on; (C) Scene containing green cube (2), red cube (5), black cylinder (3) and blue triangular prism (4)}
  \label{fig:vae_lighting_multiobj}
\end{figure}

Subsequently, the strength of the method was assessed in different lighting conditions.
We usually kept the experimental space light turned on during the two VAEs training.
In the test phase, the room light was turned off and a table light was turned on instead, for creating a different lighting condition.
The images in Figs.~\ref{fig:vae_lighting_multiobj} (A) and ~\ref{fig:vae_lighting_multiobj} (B) on the left-hand side are raw images captured by the physical Kinect.
The brightness levels of the captured images were controlled by Kinect's auto-brightness functionality, but we saw that a shadow from the green cube was changed between different conditions.
As observed from the right side of Figs.~\ref{fig:vae_lighting_multiobj} (A) and ~\ref{fig:vae_lighting_multiobj} (B), in both cases the VAE2 learned to generate very similar pseudo-synthetic images regardless of the lighting differences.

In the second set of experiments, we evaluated the validity of our method against the presence of multiple distractor objects.
The VAE2 was trained with only a single green cube object and it was subjected to the newly captured images with multiple objects without any further re-training.
As shown in the right side of Fig.~\ref{fig:vae_lighting_multiobj} (C), the VAE2 continued to successfully generate pseudo-synthetic images with only the green cube while completely ignoring the other objects in the same scene.
Therefore, this selectivity is quite useful for detecting the position of target objects even in the presence of numerous distractor objects of varying colors and shapes.

\begin{figure}[thpb]
  \centering
  \includegraphics[width=1.0\linewidth]{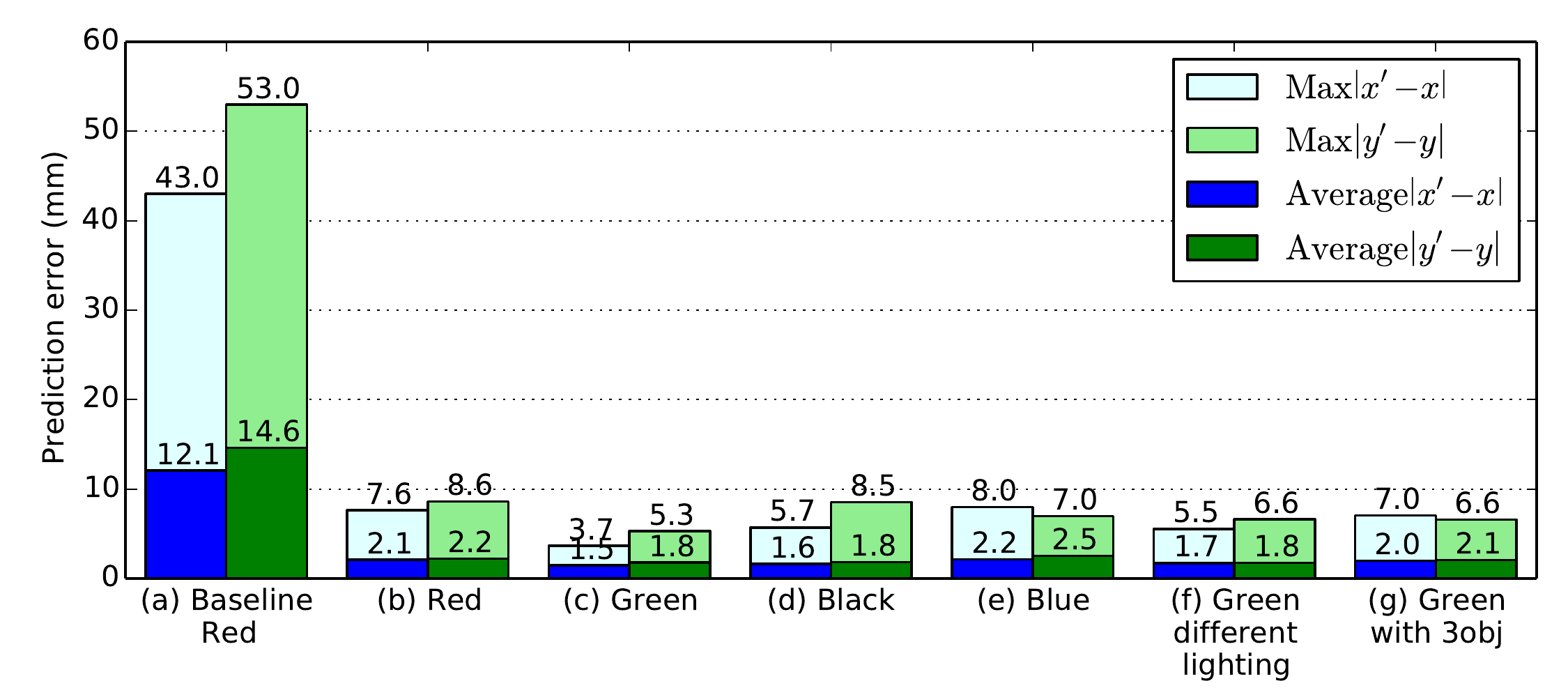}
  \caption{Experimental results for prediction errors: (a) Baseline for red cube; (b) red cube; (c) green cube; (d) black cylinder; (e) blue triangular prism; (f) green cube under different lighting condition; and (g) green cube with three other objects}
  \label{fig:precision_simple}
\end{figure}

Upon successfully learning to generate common images with the VAEs, the CNN was trained to detect object positions using $4131$ VAE1 outputs from the synthetic images generated in Gazebo.
Figure~\ref{fig:precision_simple} shows the experimental results of prediction errors for the above cases.
Compared to the baseline results shown in Fig.~\ref{fig:precision_simple} (a), our method (Fig.~\ref{fig:precision_simple} (b)) showed a considerable reduction in prediction errors.
Our method was successfully applied to differently shaped objects (Figs.~\ref{fig:precision_simple} (b)-(e)), and performed well in different lighting conditions and with other objects present (Figs.~\ref{fig:precision_simple} (f)-(g)).

\subsection{Evaluation on complex textured household objects}
\label{sec:complex_obj}

In addition, our method was applied to more complex textured objects (6)-(10) shown in Fig.~\ref{fig:natural_obj}.
We tried to detect the position of object (6) from the scenes where three objects (6)-(8) were located on a white styrofoam (experiments (h) and (i)).
As a baseline, we evaluated the outputs of the CNN trained directly by being given $54$ real images (experiment (h)).
In the real images, object (6) was located at \SI{5}{\centi\metre} grid positions with objects (7) and (8) which were located randomly on a white background.
We used $4131$ synthetic images in which a simplified orange cylinder was located at \SI{5}{\milli\metre} grid positions in Gazebo for training the VAE1 and the $54$ real images of three objects ((6)-(8)).
Then, a corresponding $54$ synthetic orange cylinder images were used for training the VAE2 (experiment (i)).

\begin{figure}[thpb]
  \centering
  \includegraphics[width=0.98\linewidth]{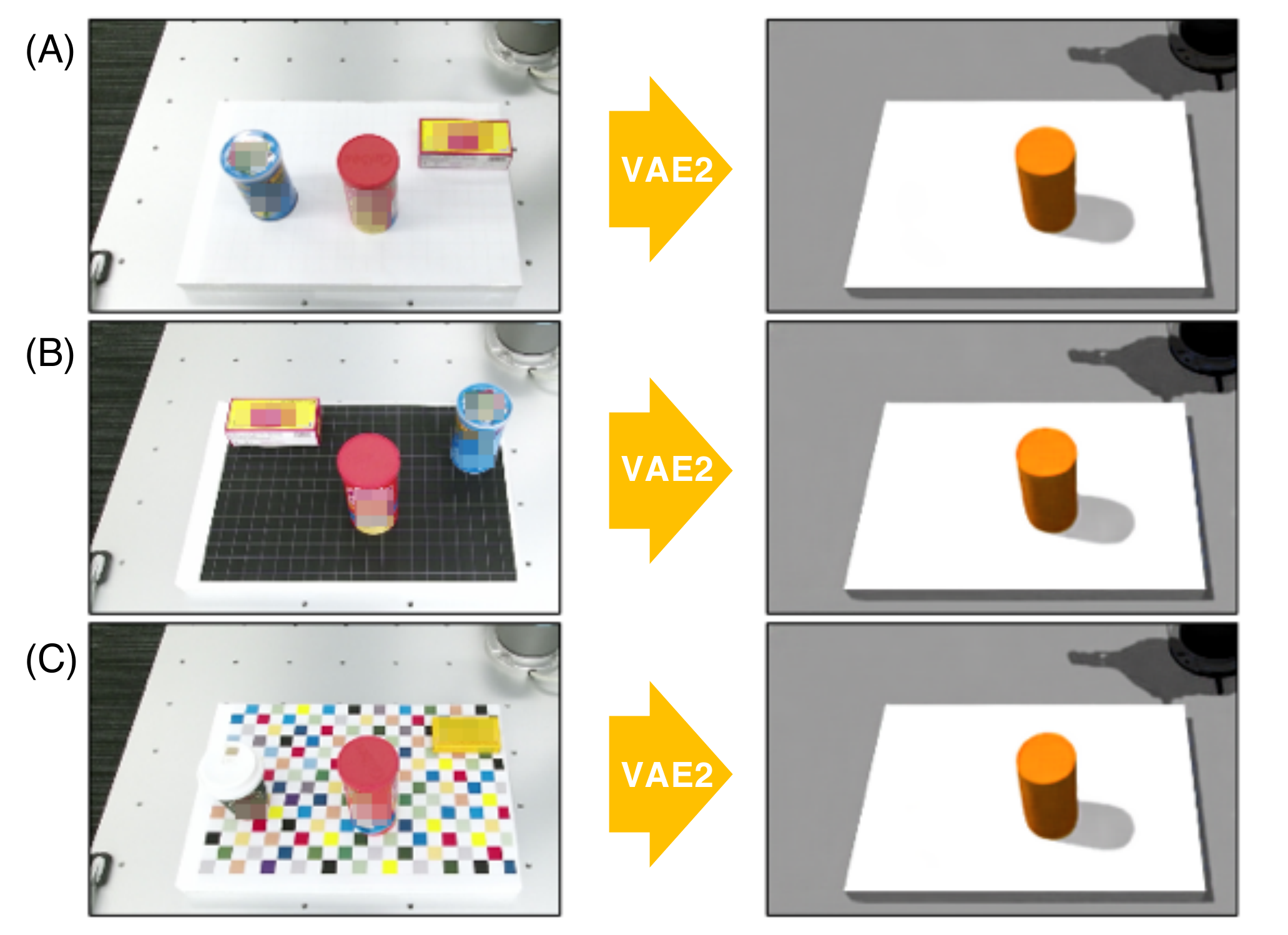}
  \caption{Images generated by VAE2 in more complex textured object cases: (A) objects (6), (7), and (8) on a white background; (B) objects (6), (7), and (8) on a black background (C) object (6), (9), and (10) on a colorful checkered background}
  \label{fig:vae_potato}
\end{figure}

Subsequently, we assessed its performance against unseen backgrounds and different distractor objects without any retraining in the test phase.
Black paper and colorful checkered paper were located on the white styrofoam as unseen backgrounds (experiments (j) and (k)).
We tried unseen object combinations with (6), (9), and (10) on the unseen black and checkered backgrounds (experiments (l) and (m)).
We also tested when there was only a single object (6) on the unseen checkered background (experiment (n)).

Figure~\ref{fig:vae_potato} shows outputs of the VAE2 trained with real images of (6), (7), and (8) on a white background.
The VAE2 could extract the object (6) in real images and reconstruct corresponding simplified orange cylinder shapes, even if the there were changes in backgrounds and other distractors.

\begin{figure}[thpb]
  \centering
  \includegraphics[width=1.0\linewidth]{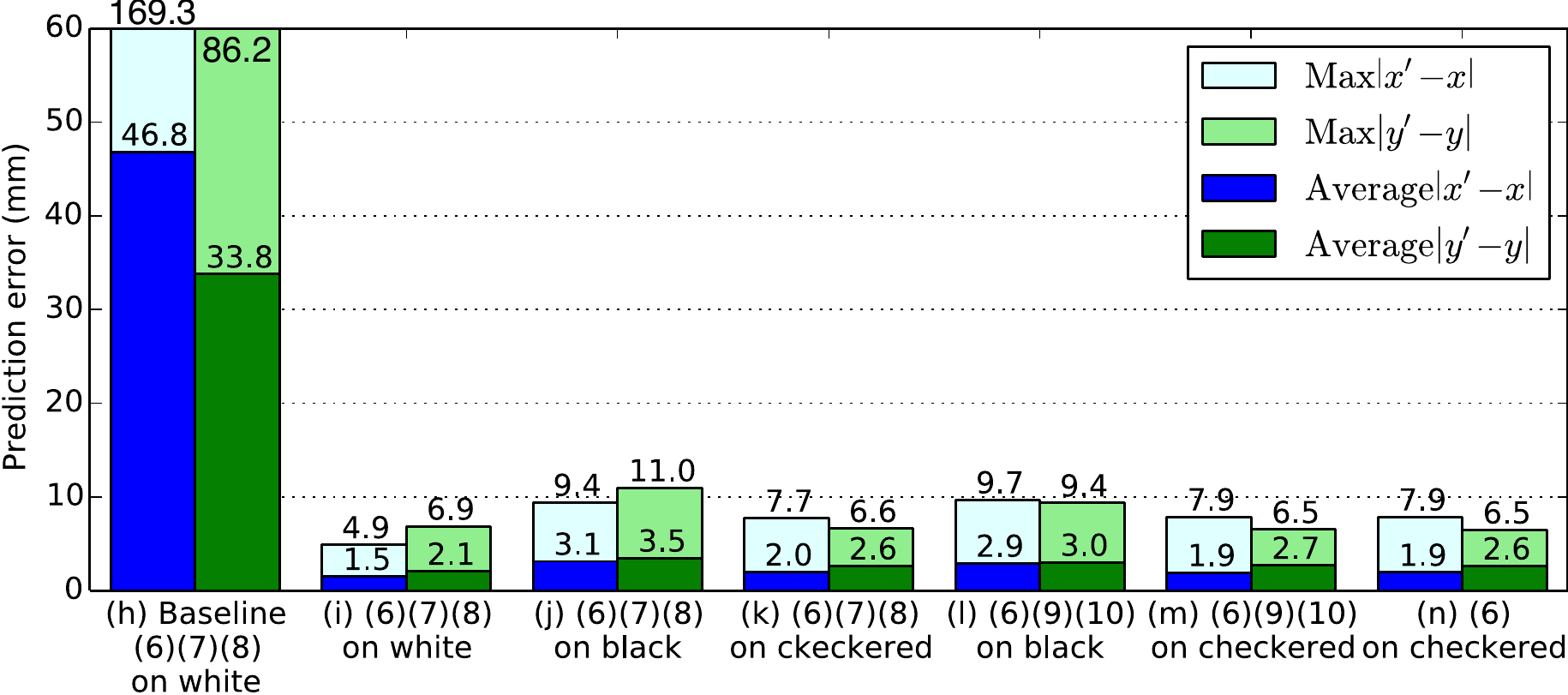}
  \caption{Experimental results for prediction errors: (h) Baseline with objects (6), (7), and (8) on white background; (i) objects (6), (7), and (8) on white background; (j) objects (6), (7), and (8) on black background; (k) objects (6), (7), and (8) on checkered background; (l) objects (6), (9), and (10) on black background; (m) objects (6), (9), and (10) on checkered background; and (n) Single object (6) on checkered background}
  \label{fig:precision_potato}
\end{figure}

Figure~\ref{fig:precision_potato} shows the experimental results of prediction errors for experiments (h)-(n).
Since the object colors were complex, the baseline result (Fig.~\ref{fig:precision_potato} (h)) was considerably degraded compared to the simple color case in Fig.~\ref{fig:precision_simple} (a).
On the other hand, our method could suppress the degradation on prediction errors shown in Fig.~\ref{fig:precision_potato} (i).
We had an accuracy of \SI{1.5}{\milli\metre} to \SI{3.5}{\milli\metre} on average on different backgrounds and with different object combinations.

Finally, we tested for precise position detection in a robotic ``pick-and-place'' task, as shown in the video
\linebreak
(\url{https://youtu.be/30vji7nJibA}).

\section{CONCLUSION}
\label{sec:conclusion}
We presented a transfer learning method using two VAEs to detect object positions precisely using only a significantly limited dataset of real images while leveraging a large dataset of synthetically generated images.
Our method performed solidly in different lighting conditions, with other objects present, and on different backgrounds.
It achieved accuracy of \SI{1.5}{\milli\metre} to \SI{3.5}{\milli\metre} on average.
We also demonstrated its efficiency in a real-world robotic application, like ``pick-and-place'' task.

\newpage

\bibliographystyle{IEEEbib}
\bibliography{refs}

\end{document}